%% file: main.tex
\documentclass[10pt,twocolumn,letterpaper]{article}

\usepackage[pagenumbers]{cvpr} 

\usepackage{colortbl}
\usepackage{multirow}
\usepackage{amsmath}
\usepackage{graphicx}
\usepackage{subcaption}
\usepackage[accsupp]{axessibility}

\definecolor{cvprblue}{rgb}{0.21,0.49,0.74}
\usepackage[pagebackref,breaklinks,colorlinks,allcolors=cvprblue]{hyperref}

\def\paperID{18121} 
\def\confName{CVPR}
\def\confYear{2025}

\title{RoboPEPP: Vision-Based \underline{Robo}t \underline{P}ose and Joint Angle Estimation \\ through \underline{E}mbedding \underline{P}redictive \underline{P}re-Training}

\author{%
  Raktim Gautam Goswami\textsuperscript{1}\thanks{Corresponding author: {\tt\small rgg9769@nyu.edu}. This paper is supported in part by the Army Research Office under grant number W911NF-21-1-0155 and by the New York University Abu Dhabi (NYUAD) Center for Artificial Intelligence and Robotics, funded by Tamkeen under the NYUAD Research Institute Award CG010.} 
  \quad Prashanth Krishnamurthy\textsuperscript{1} 
  \quad Yann LeCun\textsuperscript{2,3} 
  \quad Farshad Khorrami\textsuperscript{1} \\ [2mm]
  \textsuperscript{1}New York University Tandon School of Engineering \\
  \textsuperscript{2}New York University Courant Institute of Mathematical Sciences \quad
  \textsuperscript{3}Meta-FAIR
}

\begin{document}
\maketitle
\vspace*{-1cm}
\input{sec/0_abstract}    
\input{sec/1_intro}
\input{sec/2_related_works}
\input{sec/3_methodology}
\input{sec/4_experiments}
\input{sec/5_conclusion}

{
    \small
    \bibliographystyle{ieeenat_fullname}
    \bibliography{main}
}


\clearpage
\renewcommand{\thesection}{A\arabic{section}}
\renewcommand{\thefigure}{A\arabic{figure}}
\renewcommand{\thetable}{A\arabic{table}}
\setcounter{section}{0}  
\setcounter{figure}{0}  
\setcounter{table}{0}  
\section*{Supplementary Material}
\section{Encoder and Predictor Architectures}
As described in Sec. 3.1, we use Vision Transformer (ViT)~\cite{vit} architectures for both the encoder and predictor, similar to~\cite{jepa}. The input image, originally sized at $640 \times 480$ pixels, is cropped based on the region of interest, resized to obtain $224$ pixels along its longer side, and padded to yield a $224 \times 224$ resolution. A convolutional layer with a kernel size of $16$ and a stride of $16$ serves as the patch embedding layer, converting the image into $L$ patches of size $16 \times 16$ each with a channel dimension of $d=768$. These patches are flattened, and learnable positional embeddings, initialized as 2D sinusoidal functions, are added to the patches. The combined representations are then passed through $12$ transformer blocks. Each block contains multi-headed self-attention with $12$ heads, drop-path regularization~\cite{drop_path}, layer normalization~\cite{layer_norm}, and a multi-layer perceptron (MLP). The output of the final transformer block undergoes another layer normalization step, resulting in the encoder output $w_j \in \mathbb{R}^{768}$ for $j \in \{1, \dots, L\}$.

During evaluation, for the image of size $224 \times 224$ and a patch size of $16 \times 16$, the number of patches is computed as 
\begin{align}
L = M = \frac{224}{16} \times \frac{224}{16} = 14 \times 14 = 196.
\end{align}
However, during training, only the unmasked patches are considered, so $L < M$, i.e., $L < 196$.

The predictor takes the encoder output and reduces the embedding dimension of the patches from $768$ to $384$ using a linear layer. It also adds positional embeddings, similar to the encoder. During training, the $L (< M)$ embeddings corresponding to the unmasked patches and $(M - L)$ learnable mask tokens are concatenated to represent all patches of the original image, including the masked ones. These embeddings are then processed through $12$ transformer blocks. The final output's dimension is increased to $768$ to match the encoder's output dimension, resulting in the predictor output $v_i$ for $i \in \{1, \dots, M\}$.

The target backbone uses the same architecture as the encoder but directly operates on all $M = 196$ patches during training. It produces outputs $\bar{v}_i$ for $i \in \{1, \dots, M\}$. As outlined in the manuscript, during embedding predictive pre-training, an $L_1$ loss between $v_i$ and $\bar{v}_i$ is used to update the weights of the encoder and predictor. Following~\cite{jepa}, the target backbone is updated using an exponential moving average of the encoder’s weights.

\section{Training Settings}
\noindent\textbf{Embedding Predictive Pre-Training}:
The AdamW optimizer~\cite{adamw} with an initial learning rate of $10^{-4}$ is used for embedding predictive pre-training. The learning rate is linearly increased to $10^{-3}$ over the first 10 epochs and subsequently decreased to $10^{-6}$ using a cosine annealing scheduler. The network is pre-trained for a total of 200 epochs with a batch size of 320. Weight decay is linearly increased from $0.04$ to $0.4$ during pre-training. For the exponential moving average (EMA) update of the target backbone's weights, a momentum value of $0.996$ is used, which is linearly increased to $1.0$ over the training process.

\begin{figure*}
    \centering
    \includegraphics[width=1\linewidth]{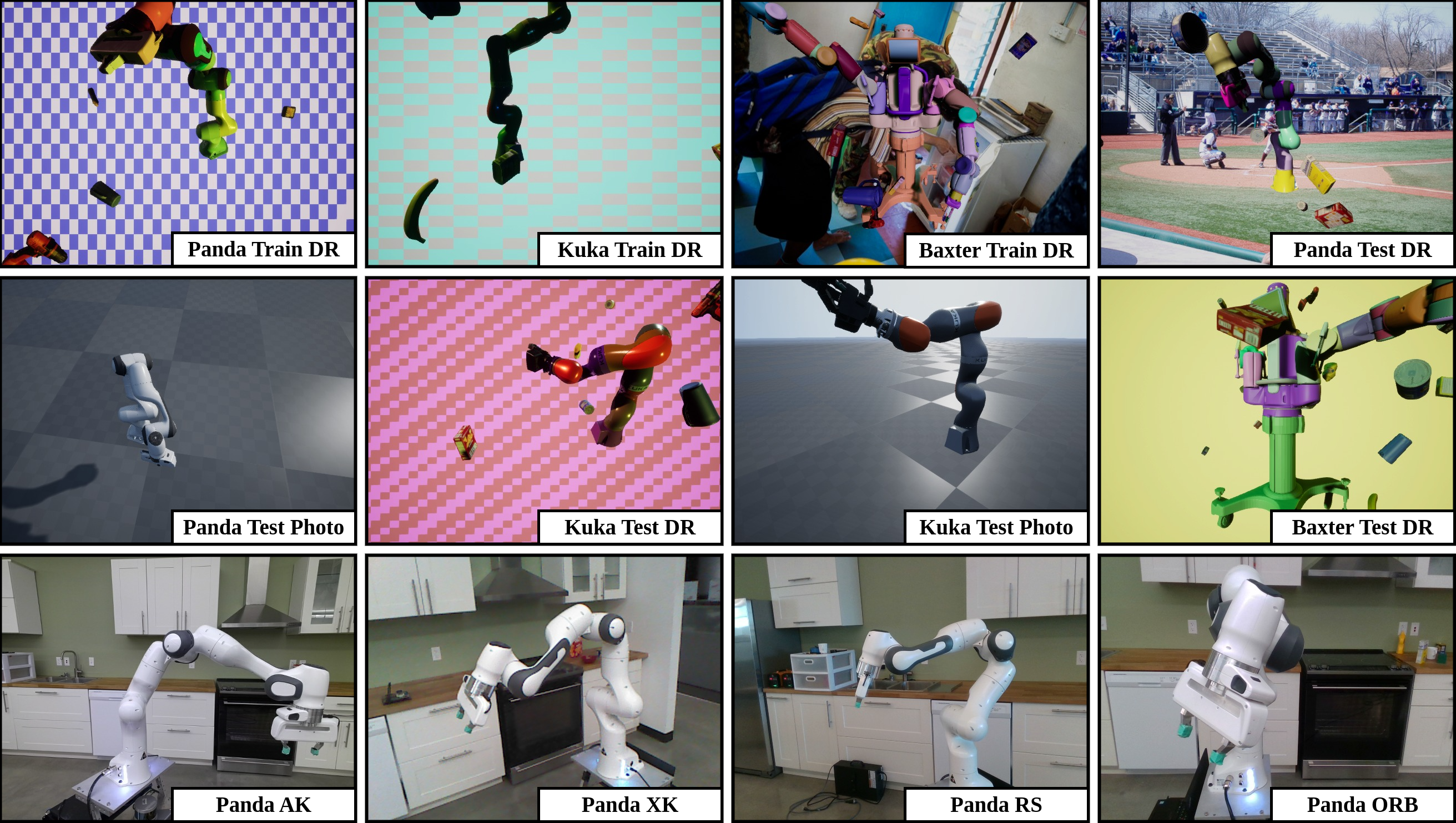}
    \caption{Example images from each of the training and test sequences from the DREAM dataset~\cite{dream}.}
    \label{fig:supp_dataset}
    \vspace*{-0.2cm}
\end{figure*}

\noindent\textbf{Keypoint Detection and Joint Angle Estimation}:
As detailed in the manuscript, the pre-trained encoder-predictor pair is fine-tuned along with the Keypoint Net and Joint Net. An AdamW optimizer~\cite{adamw} is used with an initial learning rate of $10^{-4}$, which is decreased to $10^{-8}$ using a cosine annealing scheduler. The network is trained for a total of 200 epochs with a batch size of 140.

\noindent\textbf{Sim-to-Real Self-Supervised Training}:
To bridge the sim-to-real gap, the trained networks are fine-tuned on real datasets with self-supervised training, as described in Sec. 3.3. An AdamW optimizer~\cite{adamw} is used with learning rates of $10^{-7}$ for the encoder and predictor and $10^{-5}$ for the Joint Network. The learning rate for the Keypoint Network is set close to zero to prevent model collapse. 
We observed that prioritizing Joint Network updates over the Keypoint Network yielded the best results. On the Panda RS dataset, for example, using a learning rate of $10^{-10}$ for the Keypoint Network and $10^{-5}$ for the Joint Network improved ADD AUC from 70.4 to 80.5. In contrast, reversing or equalizing the rates resulted in lower performance (71.9 and 76.4, respectively), highlighting the robustness of the keypoints for guiding the Joint Network. Further, gradients originating from the Keypoint Network's output are scaled by $10^{-10}$ to prevent conflicting updates to the encoder-predictor.
These learning rates are all decreased by a factor of $10^8$ over the training process. 
Models are fine-tuned separately for each real-world dataset for 10 epochs with a batch size of 64.

\section{Region of Interest Detection}
We utilize the pre-trained GroundingDINO~\cite{grounding_dino} object detection model to identify the region of interest, as described in Sec. 3.3. GroundingDINO is a highly accurate open-set object detector that accepts an (image, text) pair as input and outputs bounding boxes corresponding to regions of the image described by the text query. 
For all real and photo-realistic test datasets, we use the text query ``robotic arm." However, for the Panda, Kuka, and Baxter domain-randomized datasets, we use the query ``robot" because these images often contain multiple objects, some of which resemble arms and can confuse the detection model. All other parameters of GroundingDINO are left at their default values.
To address scenarios where only a portion of the robot is detected, we expand all the detected bounding boxes, especially for real datasets. Increasing all the bounding box sizes by 100 pixels on all sides generally yields robust robot pose estimation results. However, some fine-tuning of this parameter may be necessary for optimal performance depending on the specific dataset. Nonetheless, high performance is obtained even without fine-tuning.

\begin{table*}[h!]
\centering
\footnotesize
\setlength{\tabcolsep}{4.5pt} 
\renewcommand{\arraystretch}{1.3} 

\begin{tabular}{l | c c | c c c c | c}
\hline
 & \multirow{2}{*}{\shortstack{\textbf{Known} \\ \textbf{Joint Angles}}} & \multirow{2}{*}{\shortstack{\textbf{Known} \\ \textbf{Bounding Box}}} & \multicolumn{4}{c|}{\textbf{Real-World Sequences}} & \\
 &  &  & \textbf{Panda AK} & \textbf{Panda XK} & \textbf{Panda RS} & \textbf{Panda ORB} & \textbf{Average} \\
\hline
\textbf{DREAM-F} & \cellcolor{red!20}\textbf{Yes} & \cellcolor{green!20}\textbf{No} & 11413 & 491911 & 2077 & 95319 & 150180 \\
\textbf{DREAM-Q} & \cellcolor{red!20}\textbf{Yes} & \cellcolor{green!20}\textbf{No} & 78089 & 54178 & 27 & 64248 & 49136 \\
\textbf{DREAM-H} & \cellcolor{red!20}\textbf{Yes} & \cellcolor{green!20}\textbf{No} & 57 & 7382 & 24 & 25685 & 8287 \\
\hline
\textbf{HPE} & \cellcolor{green!20}\textbf{No} & \cellcolor{red!20}\textbf{Yes} & 19 & 24 & 25 & 25 & 23 \\
\hline
\textbf{RoboPose} & \cellcolor{green!20}\textbf{No} & \cellcolor{green!20}\textbf{No} & 34 & \cellcolor{gray!20}\textbf{22} & 26 & 30 & 28 \\
\textbf{HPE*} & \cellcolor{green!20}\textbf{No} & \cellcolor{green!20}\textbf{No} &  46&  -&  61&  52&  53\\
\textbf{RoboPEPP (Ours)} & \cellcolor{green!20}\textbf{No} & \cellcolor{green!20}\textbf{No} & \cellcolor{gray!20}\textbf{29} & \cellcolor{gray!20}\textbf{22} & \cellcolor{gray!20}\textbf{23} & \cellcolor{gray!20}\textbf{27} & \cellcolor{gray!20}\textbf{26} \\
\hline
\end{tabular}

\caption{Comparison of robot pose estimation using mean ADD (in millimeters), with lower a value signifying better performance. The best values among methods that use unknown joint angles and unknown bounding boxes during evaluation are bolded. HPE$^*$ denotes HPE~\cite{hpe} evaluated with the same off-the-shelf bounding box detector as RoboPEPP. HPE$^*$ was not evaluated on Panda XK since corresponding model weights were unavailable.}
\label{tab:supp_mean_ADD}
\end{table*}

\section{Dataset Details}
We evaluate our method on the DREAM dataset~\cite{dream}, which includes sequences from three robots: Franka Emika Panda (Panda), Kuka iiwa7 (Kuka), and Rethink Robotics Baxter (Baxter). The dataset provides training and testing sequences in both synthetic and real-world settings, as detailed in \Cref{tab:supp_dataset}. The synthetic data, created in Unreal Engine 4, comprises domain-randomized (DR) and photo-realistic (Photo) sequences. For real-world data, sequences of the Panda robot were captured using Microsoft Azure Kinect (AK), Xbox 360 Kinect (XK), and Intel RealSense D415 (RS) cameras, with the cameras positioned at fixed locations. Additionally, the Panda ORB dataset was collected using a RealSense camera but with varying camera placements. Example images from each dataset sequence are illustrated in \cref{fig:supp_dataset}.

\begin{table}
\centering
\renewcommand{\arraystretch}{1.2}
\begin{tabular}{c |l |c |c }
\hline
\cellcolor{gray!10} & \cellcolor{gray!10}\textbf{Dataset} & \cellcolor{gray!10}\textbf{Real} & \cellcolor{gray!10}\textbf{\# Images} \\
 \hline
 \multirow{3}{*}{\rotatebox{90}{\textbf{Training}}} & Panda Train DR & $\times$ & 104972 \\
 & Kuka Train DR & $\times$ & 104977 \\
 & Baxter Train DR & $\times$ & 104982 \\
 \hline
\multirow{9}{*}{\rotatebox{90}{\textbf{Testing}}} & Panda Photo & $\times$ & 5997 \\
 & Panda DR & $\times$ & 5998 \\
 & Panda AK & \checkmark & 6369 \\
 & Panda XK & \checkmark & 4966 \\
 & Panda RS & \checkmark & 5944 \\
 & Panda ORB & \checkmark & 32315 \\
 & Kuka Photo & $\times$ & 5999 \\
 & Kuka DR & $\times$ & 5997 \\
 & Baxter DR & $\times$ & 5982 \\
 \hline

\end{tabular}
\caption{Number of images in each sequence of the dataset.}
\label{tab:supp_dataset}

\end{table}

\section{Additional Results}
\subsection{Mean ADD}
In \Cref{tab:supp_mean_ADD}, we present the mean ADD (Average Distance) values (ADD defined in Sec.~4.2.1) on the Panda real-world datasets. Consistent with Table~2, we compare our method, RoboPEPP, against DREAM~\cite{dream}, RoboPose~\cite{robopose}, HPE~\cite{hpe}, and HPE$^*$ (HPE using our bounding box detection strategy). RoboPEPP achieves the lowest mean ADD across all real-world data sequences among methods that operate with unknown joint angles and bounding boxes. DREAM~\cite{dream}, which detects 2D keypoints and employs them in a PnP solver to estimate the robot pose, is highly sensitive to keypoint detection errors. Even a single incorrectly detected keypoint can cause DREAM to fail in pose estimation, leading to high ADD.

\subsection{Ablation: Occlusion Robustness}
\begin{figure}
    \centering
    \includegraphics[width=1\linewidth]{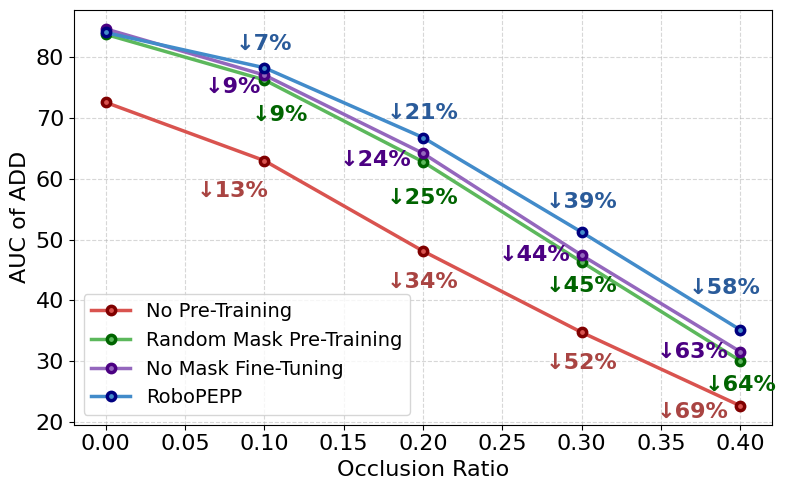}
    \caption{AUC comparison of the distance metric under varying occlusion levels, evaluated on the dataset in Sec. 4.2.3. Percentages next to the lines indicate the relative drop in each method’s performance compared to their performance without occlusions.}
    \label{fig:supp_occ_res}
\end{figure}

In this section, we evaluate the methods from the \textit{Embedding Predictive Pre-Training} ablation studies (Sec.~4.3) on the occlusion dataset described in Sec.~4.2.3. Specifically, we compare the following models: (1) a version of RoboPEPP without pre-training, (2) a version pre-trained with random masking instead of joint-specific masking, (3) the standard RoboPEPP (pre-trained with joint masking), and (4) a model pre-trained with joint masking but fine-tuned without masking during the encoder-predictor fine-tuning phase. 
As shown in \cref{fig:supp_occ_res}, and similar to Fig.~6, we plot the AUC of the ADD metric against the occlusion ratio. Additionally, the percentage decrease in AUC relative to the performance without occlusion is annotated on the plot. Among the methods, RoboPEPP achieves the best performance across all occlusion ratios. While the framework with random-masking-based pre-training and the one fine-tuned without masking achieve performance comparable to RoboPEPP under zero occlusion, their performances degrade more rapidly as the occlusion ratio increases.

\subsection{Ablation: Joint Net and Keypoint Net}
\begin{figure}[h!]
    \centering
    \includegraphics[width=0.8\linewidth, trim={0 0 0 0}, clip]{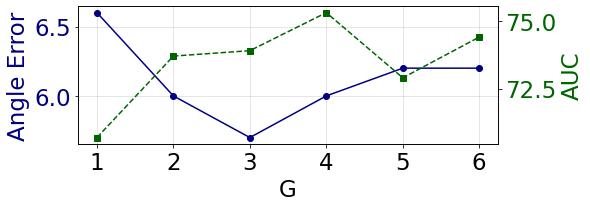}
    \caption{Average AUC and Joint Angle errors on the Panda dataset for different values of Joint Net's G.}
    \label{fig:supp_ablation_angle}
    \vspace*{-1cm}
\end{figure}

Fig. \ref{fig:supp_ablation_angle} and Table \ref{tab:supp_ablation_channel} analyze the Joint Net's $G$ value (number of refinement steps) and Keypoint Net's hidden channel dimension, respectively, with $G$=4 and 256 channels yielding the best overall results.

\begin{table}[h!]
    \centering
    \begin{tabular}{c|c}
        \hline
        \cellcolor{gray!30}\textbf{Channel Dim.} & \cellcolor{gray!30}\textbf{AUC}  \\ \hline
        128 & 69.3 \\ 
        256 & \cellcolor{gray!20}\textbf{75.5} \\ 
        512 & 74.4 \\ 
    \end{tabular}
    \caption{Average AUC on the Panda dataset for different values of the Keypoint Net's hidden channel dimension.}
    \label{tab:supp_ablation_channel}
    \vspace*{-0.14in}
\end{table}

\section{Additional Qualitative Comparison}
\begin{figure*}
    \centering
    \includegraphics[width=1\linewidth]{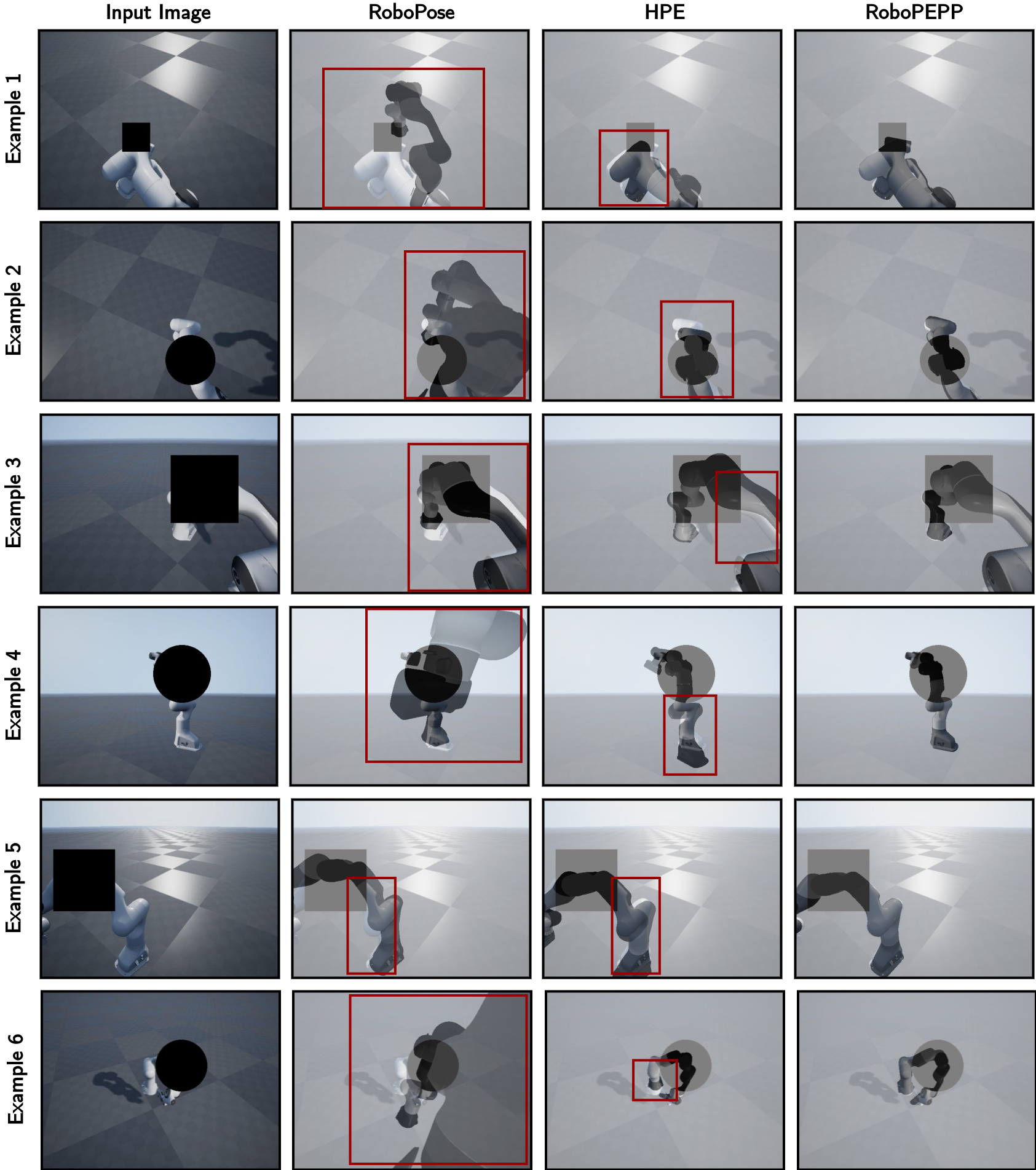}
    \caption{\textbf{Qualitative Comparison on Occlusion dataset}: Predicted poses and joint angles are used to generate a mesh overlaid on the original image, where closer alignment indicates greater accuracy. Highlighted rectangles indicate regions where other methods’ meshes misalign, while RoboPEPP achieves high precision.}
    \label{fig:supp_qual_occ_comp}
\end{figure*}

\begin{figure*}
    \centering
    \includegraphics[width=0.9\linewidth]{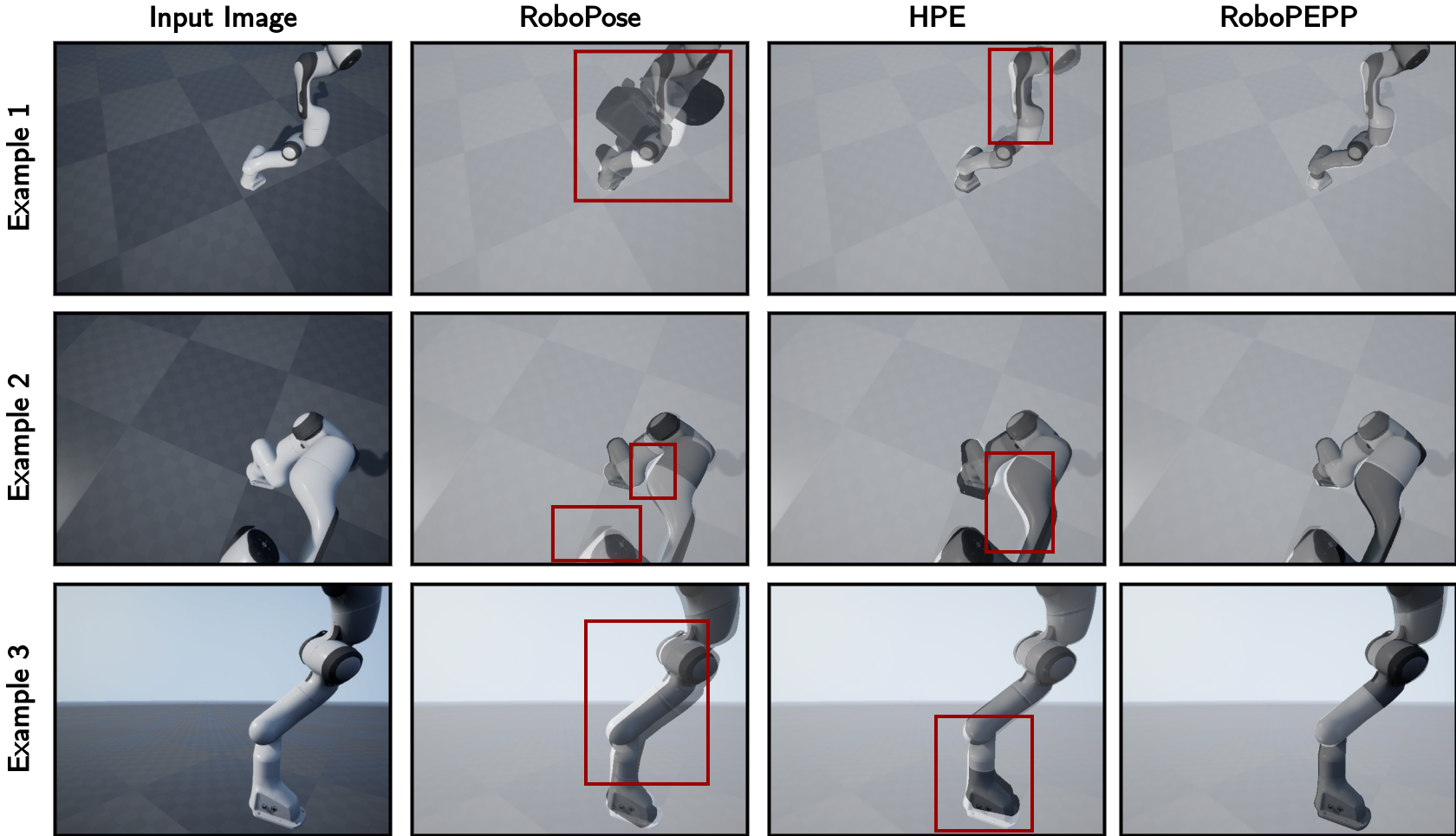}
    \caption{\textbf{Qualitative Comparison on Panda Photo dataset}: Predicted poses and joint angles are used to generate a mesh overlaid on the original image, where closer alignment indicates greater accuracy. Highlighted rectangles indicate regions where other methods’ meshes misalign, while RoboPEPP achieves high precision.}
    \label{fig:supp_qual_photo_comp}
\end{figure*}

\begin{figure*}
    \centering
    \includegraphics[width=0.9\linewidth]{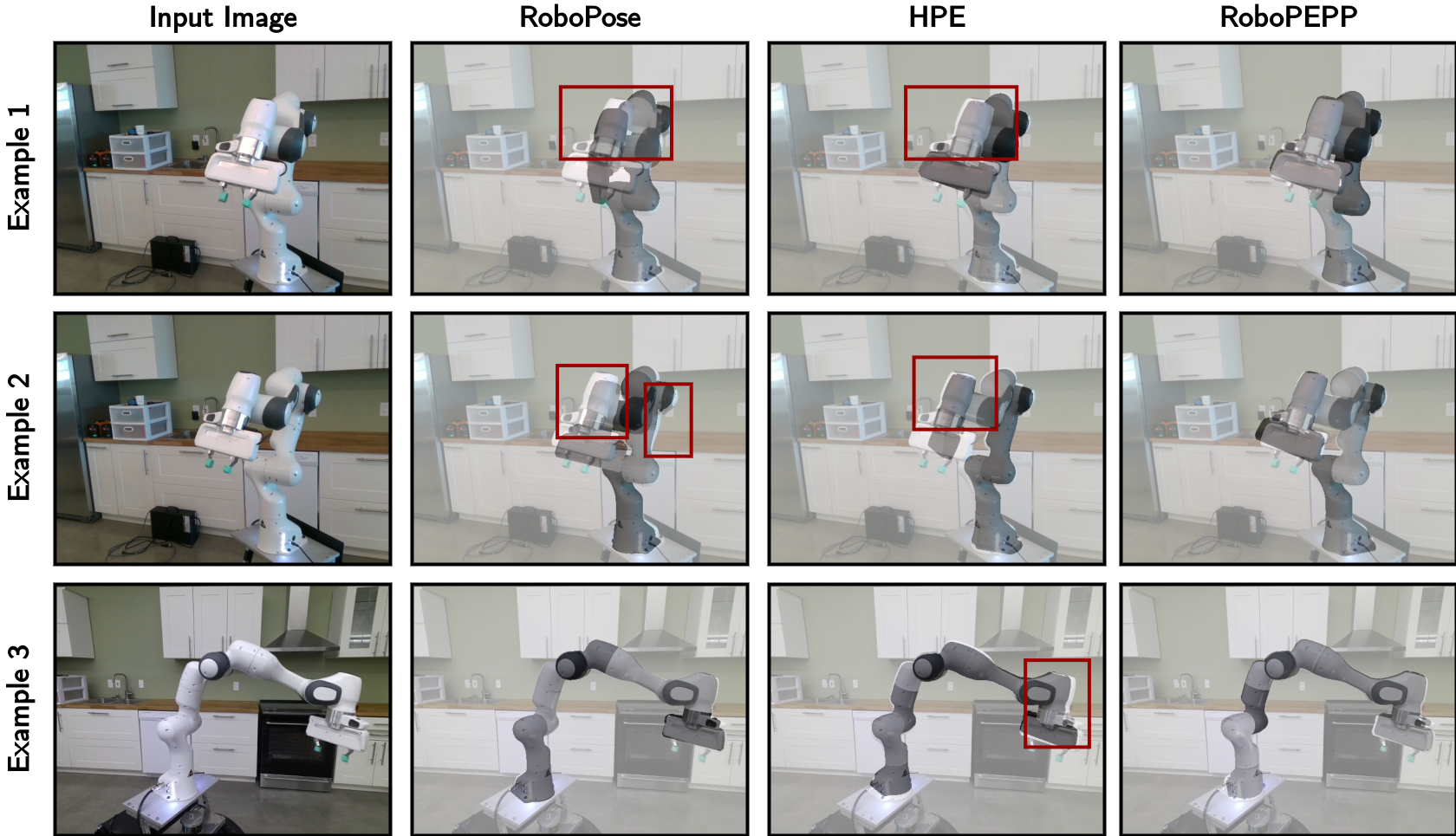}
    \caption{\textbf{Qualitative Comparison on Panda RS (Example 1 and 2) and Panda AK (Example 3) datasets}: Predicted poses and joint angles are used to generate a mesh overlaid on the original image, where closer alignment indicates greater accuracy. Highlighted rectangles indicate regions where other methods’ meshes misalign, while RoboPEPP achieves high precision.}
    \label{fig:supp_qual_real_comp}
\end{figure*}

\begin{figure*}
    \centering
    \includegraphics[width=1\linewidth]{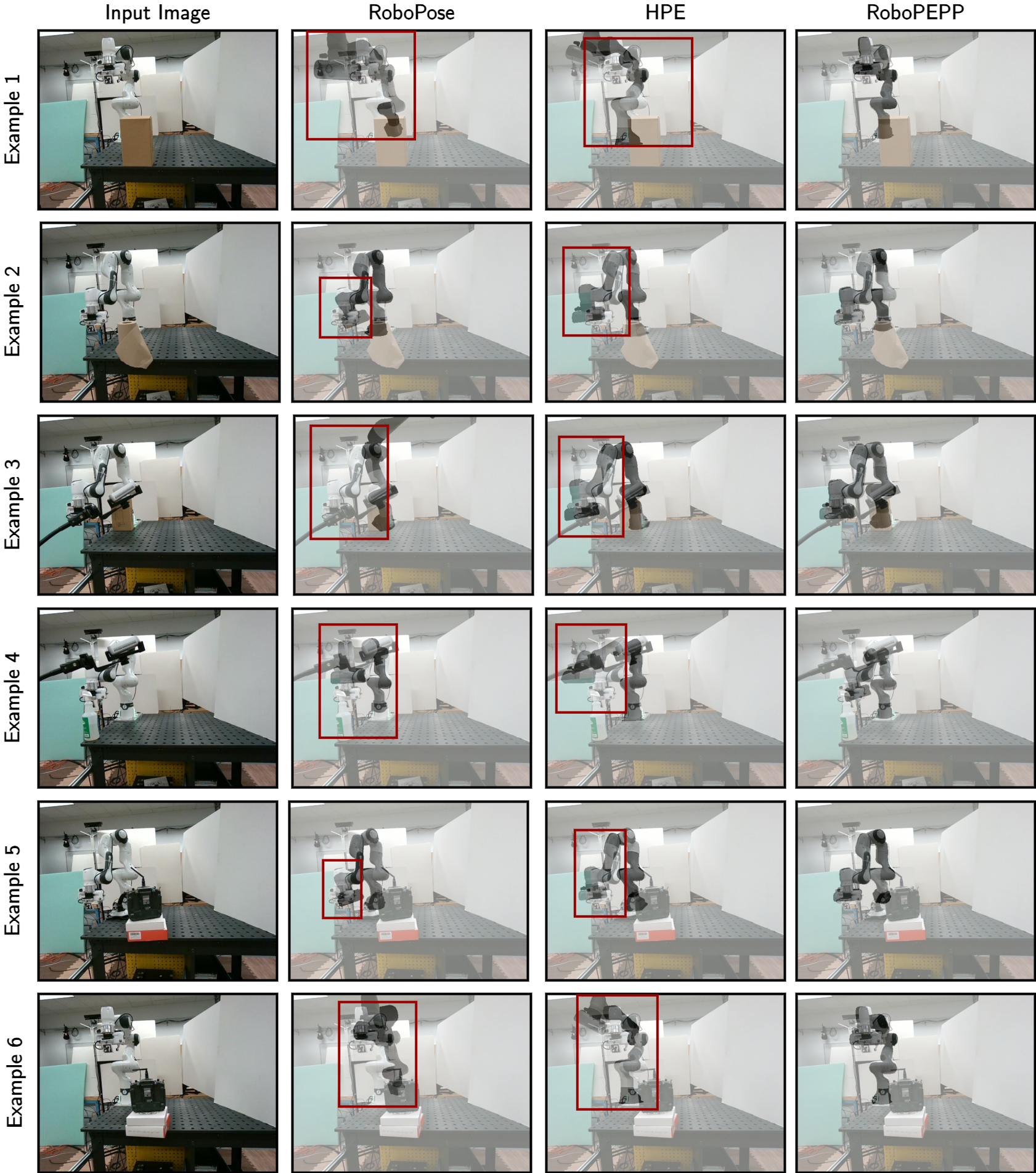}
    \caption{\textbf{Qualitative Comparison on Additional Real-World Images}: These images are collected in highly cluttered environments with robot occlusions. Predicted poses and joint angles generate a mesh overlaid on the original image, where closer alignment indicates greater accuracy. Highlighted rectangles indicate regions where other methods’ meshes misalign, while RoboPEPP achieves high precision.}
    \label{fig:supp_qual_crrl_franka_comp}
\end{figure*}

In this section, we provide additional examples of qualitative comparisons. \cref{fig:supp_qual_occ_comp} presents examples from the occlusion dataset discussed in Sec. 4.2.3. \cref{fig:supp_qual_photo_comp} shows comparisons on the Franka Photo dataset, while \cref{fig:supp_qual_real_comp} highlights results on the real-world datasets Franka RS and AK. Lastly, \cref{fig:supp_qual_crrl_franka_comp} focuses on real-world images of the Franka robot collected in the lab under highly cluttered and occluded conditions. 
For all examples, comparisons are made against RoboPose~\cite{robopose} and HPE~\cite{hpe}. Rectangles are used to emphasize areas where these methods perform poorly, while RoboPEPP shows higher accuracy.


\end{document}

%% file: sec/0_abstract.tex
\begin{abstract}
Vision-based pose estimation of articulated robots with unknown joint angles has applications in collaborative robotics and human-robot interaction tasks. 
Current frameworks use neural network encoders to extract image features and downstream layers to predict joint angles and robot pose. 
While images of robots inherently contain rich information about the robot's physical structures, existing methods often fail to leverage it fully; therefore, limiting performance under occlusions and truncations. 
To address this, we introduce RoboPEPP, a method that fuses information about the robot’s physical model into the encoder using a masking-based self-supervised embedding-predictive architecture.
Specifically, we mask the robot's joints and pre-train an encoder-predictor model to infer the joints' embeddings from surrounding unmasked regions, enhancing the encoder's understanding of the robot's physical model.
The pre-trained encoder-predictor pair, along with joint angle and keypoint prediction networks, is then fine-tuned for pose and joint angle estimation. 
Random masking of input during fine-tuning and keypoint filtering during evaluation further improves robustness. 
Our method, evaluated on several datasets, achieves the best results in robot pose and joint angle estimation while being the least sensitive to occlusions and requiring the lowest execution time. The code is available at \href{https://github.com/raktimgg/RoboPEPP}{https://github.com/raktimgg/RoboPEPP}.
\vspace*{-0.5cm}
\end{abstract}

%% file: sec/1_intro.tex
\section{Introduction}
\label{sec:intro}
Estimating the pose and joint angles of an articulated robot in the coordinate frame of an external camera is valuable for facilitating collaborative applications wherein an agent (e.g., a human or another robot) operates in a shared space with the articulated robot~\cite{hci1,hci2,hci3,multi_robot1,multi_robot2,multi_robot3}.
Traditional robot pose estimation methods assume known joint angles and capture multiple images with fiducial markers~\cite{aruco,apriltag,artag} attached to the robot's end-effector to establish 2D-3D correspondences between the image pixels and the robot’s frame.
\begin{figure}
    \centering
    \includegraphics[width=0.95\linewidth]{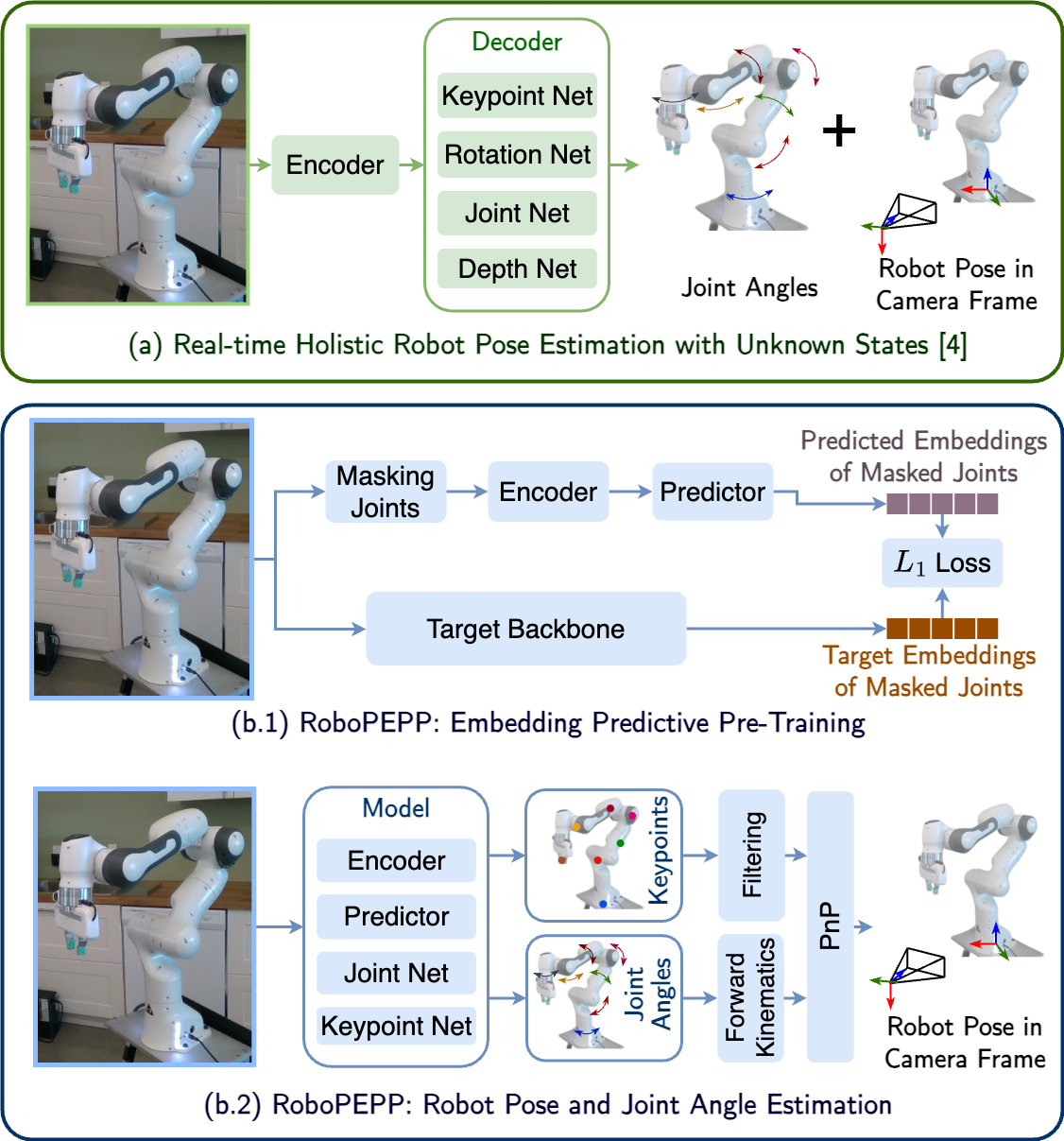}
    \caption{Comparison of an existing robot pose estimation method~\cite{hpe} with our RoboPEPP framework. RoboPEPP integrates joint masking-based pre-training (b.1) to enhance the encoder's grasp of the robot's physical model, combined with downstream networks, and keypoint filtering (b.2) to achieve high accuracy.}
    \vspace*{-0.7cm}
    \label{fig:teaser}
\end{figure}
Recent advancements in deep learning enable the prediction of 2D keypoints on robot joints from a single image~\cite{dream, ctrnet, posefusion, tian2023robot}. However, the assumption of known joint angles is not valid in many practical settings such as in collaborative robotics and human-robot interaction, where the joint angles may be unreliable or completely unknown. 
This challenge of simultaneously estimating joint angles and robot poses is particularly complex due to the high degrees of freedom in robotic systems and the infinite space of potential robot poses and joint angle configurations.
RoboPose~\cite{robopose} pioneered the field of robot pose estimation with unknown joints by using an iterative render-and-compare strategy. Later works~\cite{hpe,robokeygen} enhanced the efficiency by employing neural networks that predict joint angles and robot poses in a single feed-forward pass. While input images provide rich information about the robot's physical structures and constraints, existing methods fail to leverage this fully, resulting in low performance in challenging scenarios like occlusions and truncations (i.e., instances where only part of the robot is visible).

Recently, self-supervised learning~\cite{jepa, vjepa} has shown that embedding predictive pre-training helps encoders develop a deeper semantic understanding of images. 
Inspired by such works, we propose RoboPEPP (\cref{fig:teaser}), a robot pose estimation framework that integrates a joint-masking-based pre-training strategy to help the encoder better understand the robot's physical model. In this approach, the encoder extracts embeddings from the unmasked regions, which a predictor uses to estimate embeddings of the masked joints.
In other words, the encoder-predictor network is trained to predict the embeddings of the joints using the context around them, thus improving the network's understanding of the robot's structure.
While this pre-trained encoder supports various robotics tasks, we focus on robot pose estimation.

Following pre-training, the encoder and predictor are fine-tuned using downstream layers for joint angle prediction and 2D keypoint heatmap generation, allowing for end-to-end training. We further enhance the model's occlusion robustness by randomly masking the input while fine-tuning. 
During inference, the pixels with the highest values in the heatmaps are identified as 2D keypoints, while corresponding 3D keypoints in the robot’s frame are computed using the forward kinematics and predicted joint angles. For cases where only part of the robot is visible in the image, we apply confidence-based keypoint filtering. Finally, we use the perspective-$n$-point (PnP) algorithm~\cite{epnp} on the filtered 2D-3D correspondences to estimate the robot's pose.
In summary, our contributions are:
\begin{itemize}
    \item A robot pose and joint angle estimation framework with embedding-predictive pre-training to enhance the network’s understanding of the robot’s physical model.
    \item An efficient network for robot pose and joint angle estimation using the pre-trained encoder-predictor alongside joint angle and keypoint estimators, trained using randomly masked inputs to enhance occlusion robustness.
    \item A confidence-based keypoint filtering method to handle cases where only part of the robot is visible in the image.
    \item Extensive experiments showing RoboPEPP's superior pose estimation, joint angle prediction, occlusion robustness, and computational efficiency.
\end{itemize}

\begin{figure*}
    \centering
    \includegraphics[width=0.85\linewidth,clip=true,trim=0in 0.22in 0in 0in]{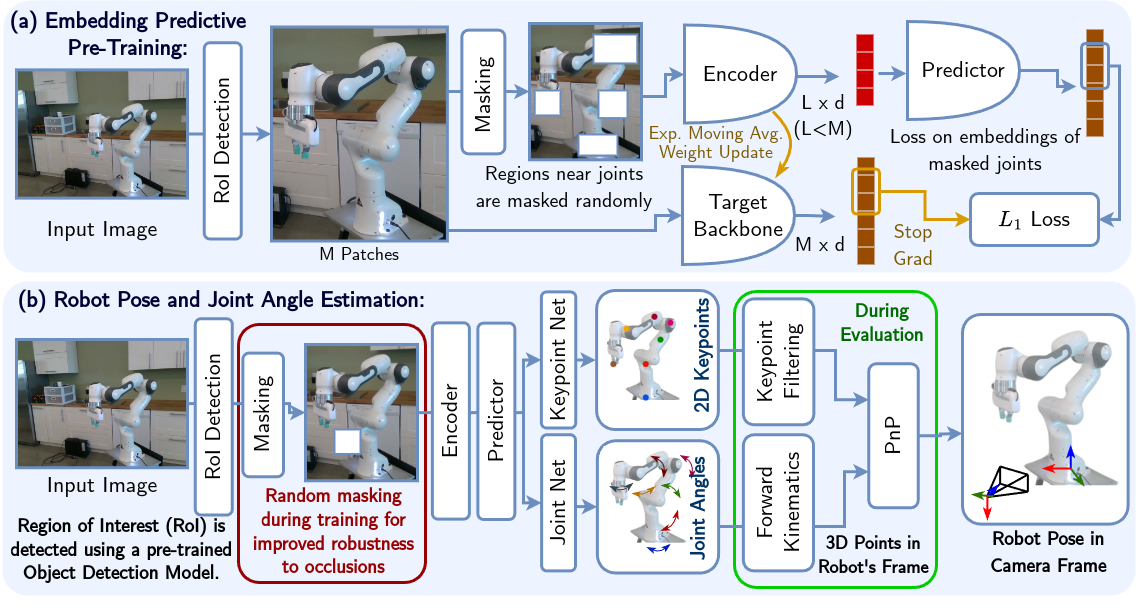}
    \caption{Overview of the RoboPEPP framework for robot pose and joint angle estimation. (a) Joint regions are masked to pre-train an encoder-predictor pair using an embedding predictive architecture. (b) The pre-trained encoder-predictor network is fine-tuned for robot pose estimation with Joint and Keypoint Prediction networks, using random masking during training to enhance occlusion robustness. During evaluation, keypoints are filtered, and a PnP algorithm estimates the robot’s pose from the filtered 2D-3D correspondences.}
    \label{fig:flow}
    \vspace*{-0.5cm}
\end{figure*}


%% file: sec/2_related_works.tex
\section{Related Work}
\label{sec:related_work}
Classical methods for robot pose estimation typically involve attaching fiducial markers, such as ArUco~\cite{aruco}, AprilTag~\cite{apriltag}, or ARTag~\cite{artag}, to the robot's end-effector to obtain easily detectable pixels in images. The corresponding 3D points in the robot's base coordinate frame are calculated using the robot's joint angles and forward kinematics. Using these correspondences and the camera intrinsics, an optimization problem is solved to find the robot-to-camera transformation~\cite{hand2eye_eg1, hand2eye_eg2, hand2eye_eg3}, referred to here as the robot pose. This, however, requires multiple sets of correspondences from images taken at different robot configurations.

To streamline this process, DREAM~\cite{dream} introduced a learning-based approach to detect multiple keypoint correspondences from a single image, estimating the pose using the PnP~\cite{epnp} algorithm. This method achieved performance comparable to classical approaches while requiring a single image. Building on this, PoseFusion~\cite{posefusion} used multi-scale feature fusion to improve keypoint prediction accuracy. 
G-SAM~\cite{gsam} further improved robustness by adding a grouping and soft-argmax module, particularly useful when only part of the robot is visible.
CTRNet~\cite{ctrnet} introduced a self-supervised sim-to-real approach, using differentiable PnP solver~\cite{differentiable_pnp} and mesh renderer~\cite{pytorch3d} to predict robot masks, which are compared against foreground segmentation for training. To incorporate information from prior frames, SGTAPose~\cite{sgtapose} employed a temporal attention framework.

These approaches, however, assume known joint angles, which is often impractical in real-world settings like collaborative robotics and human-robot interaction. 
To address this, RoboPose~\cite{robopose} estimated pose with unknown joint angles using an iterative render-and-compare approach, yielding strong results but at the cost of computational efficiency. An efficient framework~\cite{hpe} was later developed to predict joint angles and pose in a single, real-time feed-forward pass. RoboKeyGen~\cite{robokeygen} proposed another efficient method by lifting 2D keypoints to 3D using a stable diffusion model. 
While some frameworks~\cite{SPDH,3d_pose_nowcasting} used depth cameras for pose prediction, we restrict the discussion to methods using monocular RGB images, which are more widely accessible and avoid the need for specialized sensors.

Current robot pose estimation methods~\cite{robopose,hpe,robokeygen} do not fully utilize the rich features of the robot's physical model available in images. Meanwhile, self-supervised learning frameworks~\cite{ssl1,ssl2,ssl3,ssl4,jepa, vjepa} have advanced encoder training to extract robust image features with Joint-Embedding Predictive Architecture (JEPA)~\cite{jepa} using a masking-based pre-training strategy to enhance the encoder's semantic understanding of the image.
Inspired by JEPA, we pre-train an encoder-predictor pair by masking regions around the robot's joints and predicting embeddings of the masked regions based on the surrounding context, thus enhancing the encoder's understanding of the robot's physical model.
The pre-trained encoder-predictor pair is then fine-tuned along with joint and keypoint prediction networks, applying random masking during fine-tuning and confidence-based keypoint filtering at evaluation for improved robustness.

%% file: sec/3_methodology.tex
\section{Methodology}
\label{sec:methodology}
\textbf{Problem Description:}
Given a color image capturing an articulated robot with \( n \) joints, our objective is to estimate the joint angles \( \Phi \in \mathbb{R}^n \) and the robot-to-camera rigid transformation matrix \( T_R^C \in SE(3) \), with the robot frame being defined at its base.
The robot’s forward kinematics and the camera's intrinsic parameters are assumed to be known.

\noindent\textbf{Method Overview:}
Our proposed framework (\cref{fig:flow}) comprises two stages: self-supervised pre-training of an encoder-predictor network (\cref{sec:pre-training}); and fine-tuning of the pre-trained encoder-predictor alongside 2D keypoint detection and joint angle estimation networks (\cref{sec:kp_joint_estimation}).
Predicted joint angles and forward kinematics yield 3D joint coordinates, which, combined with detected 2D keypoints, are used in a PnP solver to estimate pose (\cref{sec:pose_estimation}). During evaluation, confidence-based keypoint filtering and self-supervised fine-tuning on real-world data enhance accuracy.

\subsection{Embedding Predictive Pre-Training}
\label{sec:pre-training}
Building on embedding predictive architectures~\cite{jepa, vjepa}, we employ a masking-based pre-training strategy tailored for robotic applications like pose and joint estimation. Masks are selected to occlude the regions around four randomly selected robot joints, or a random area if a joint is outside the camera's field of view. Each mask covers 15–20\% of the image with an aspect ratio between 0.75 and 1.5.

The original image consists of \( M \) patches, each sized \( 16 \times 16 \) pixels. Let \( u_i \) represent the \( i \)-th patch, where \( i \in \{ 1, 2, \dots, M \} \), and let \( \mathcal{B} \) denote the set of indices for the unmasked patches, with \( L = |\mathcal{B}| < M \). With patches \( u_j \), for \( j \in \mathcal{B} \), as the context, a Vision Transformer (VIT)~\cite{vit} encoder produces context embeddings \( w_j \in \mathbb{R}^d \) for \( j \in \mathcal{B} \). These context embeddings are then passed to a VIT-based predictor, which infers embeddings for all \( M \) patches of the original image, denoted \( v_i \in \mathbb{R}^d \) for \( i \in \{ 1, 2, \dots, M \} \). 

Meanwhile, a target backbone with the same architecture as the encoder extracts embeddings \( \bar{v}_i \in \mathbb{R}^d \) for \( i \in \{ 1, 2, \dots, M \} \) directly from the original image. 
The embeddings for the masked patches, corresponding to indices \( i \in \bar{\mathcal{B}} \) (where \( \bar{\mathcal{B}} \) denotes the set of masked patch indices), are used to compute the \( L_1 \) loss during training, given by:
\vspace*{-0.2cm}
\begin{align}
   \mathcal{L}_{\text{pre-train}} = \frac{1}{|\bar{\mathcal{B}}|} \sum_{i \in \bar{\mathcal{B}}} \|v_i - \bar{v}_i\|_1 .
\end{align}
Backpropagating through the encoder-predictor network and target backbone simultaneously risks trivial solutions, like constant predictions across all networks. To avoid this, we follow~\cite{jepa}, backpropagating only through the encoder-predictor branch and updating the target backbone with an exponential moving average of the encoder's weights.

Our approach differs from JEPA~\cite{jepa} by using context-informed masking at joint locations. While JEPA learns deeper semantic representations by randomly masking the input for tasks like object detection, we focus on encoding the robot’s physical properties by specifically masking joint regions. This trains the encoder to infer the robot's joint-related information based on the surroundings, emulating a predictive understanding similar to how humans or animals deduce missing information about physical structures. 

\subsection{Keypoint Detection and Joint Angle Estimation}
\label{sec:kp_joint_estimation}
The pre-trained encoder and predictor are then fine-tuned, where they extract embeddings \( v_i \in \mathbb{R}^d \) for \( i \in \{ 1, 2, \dots, M \} \) from images, which are used by the Joint Net and Keypoint Net to predict joint angles and 2D keypoints, respectively. To further increase occlusion robustness, random masks covering up to 20\% of the image are applied during training. Consistent with \cref{sec:pre-training}, the predictor outputs all patch embeddings, including masked ones. This framework is trained using the loss functions in \cref{sec:loss}.

\begin{figure}
    \centering
    \includegraphics[width=1\linewidth]{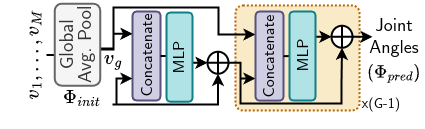}
    \caption{\textbf{Joint Net}: A global average pooling layer aggregates the patch embeddings, $v_1, \dots, v_M$, into $v_g$, which is then iteratively refined using an MLP to estimate the joint angles.}
    \label{fig:joint_net}
    \vspace*{-0.5cm}
\end{figure}

\subsubsection{Joint Net}
Using the patch embeddings, $v_i$, as input, the Joint Net predicts the angles for each of the robot's \( n \) joints. 
A global average pooling layer aggregates the patch embeddings \( v_i \) (for \( i \in \{ 1, 2, \dots, M \} \)) into a single embedding \( v_g \in \mathbb{R}^d \) to generate a global representation of the image. An iterative MLP-based approach~\cite{hpe} is then used to refine the joint angle predictions. Starting with a zero vector as the initial estimate \( \Phi_{\text{init}} \), the joint angles are iteratively updated through the MLP over \( G = 4 \) refinement steps (\cref{fig:joint_net}). The same MLP layer is used across all iterations, progressively refining the predicted joint angles \( \Phi_{\text{pred}} \) for improved accuracy.

\subsubsection{Keypoint Net}
The Keypoint Net uses the patch embeddings to predict heatmaps for each of the $k$ keypoints. The matrix \( V = [v_1, v_2, \dots, v_M]^T \in \mathbb{R}^{M \times d} \), contianing the patch embeddings, is reshaped into \( \hat{V} \) \(\in \mathbb{R}^{m\times m\times d} \), where \( m = \sqrt{M} \). With input image of \( 224 \times 224 \) pixels and a patch size of \( 16 \times 16 \) pixels, \( m = 14 \).
The Keypoint Net takes \( \hat{V} \) as input and applies four upsampling layers with output dimensions shown in Table \ref{tab:kp_net}. Each upsampling layer includes a transpose convolutional layer with a kernel size of 4, stride of 2, and one-pixel wide zero padding, followed by batch normalization, ReLU activation, and dropout. The channel dimension is gradually reduced from \( d = 768 \) to 256 across these layers. The output is then passed through a linear layer that reduces the channel dimension to \( k \), followed by a sigmoid activation to produce heatmaps $H_{pred} \in \mathbb{R}^{224\times224\times k}$.
Typically, each keypoint is defined at a joint of the robot, with an additional keypoint at the base, making \( k = n + 1 \).

\begin{table}[h!]
\centering
\small
\begin{tabular}{l|c|c}
\hline
\cellcolor{gray!20}\textbf{Layer}       & \cellcolor{gray!20}\textbf{Spatial Size}   & \cellcolor{gray!20}\textbf{Channels}    \\ \hline
Input Size           & 14 $\times$ 14 & 768              \\ \hline
Upsample 1           & 28 $\times$ 28 & 256              \\ 
Upsample 2           & 56 $\times$ 56 & 256              \\ 
Upsample 3           & 112 $\times$ 112 & 256            \\ 
Upsample 4           & 224 $\times$ 224 & 256            \\ \hline
Linear (Heatmaps)    & 224 $\times$ 224 & k              \\ 
\end{tabular}
\caption{\textbf{Layer Output Sizes in Keypoint Net}: Patch embeddings are progressively upsampled through four layers and the channel dimension is reduced to k (the number of keypoint).}
\label{tab:kp_net}
\vspace*{-0.4cm}
\end{table}


\subsubsection{Loss Functions}
\label{sec:loss}
For joint angles, we employ a mean squared error loss:
\vspace*{-0.2cm}
\begin{align}
    \mathcal{L}_{joint} = \frac{1}{n} \| \Phi_{pred} - \Phi_{gt}\|_2^2
\end{align}
where \( \Phi_{pred} \) and \( \Phi_{gt} \) represent the predicted and ground truth joint angles, respectively. To enhance training convergence, mean-variance normalization is applied to \( \Phi_{gt}\).
For keypoint detection, we utilize the focal loss~\cite{focal_loss, cornernet} ($\mathcal{L}_{kp}$):
\vspace*{-0.2cm}
\begingroup
\thinmuskip=1mu  
\medmuskip=1mu   
\thickmuskip=1mu 
\begin{align}
    \frac{-1}{\mathcal{S}} \sum_{i=1}^{\mathcal{H}} \sum_{j=1}^{\mathcal{W}} 
        \begin{cases} 
        (1 - p_{ij})^\alpha \log(p_{ij}), & \text{\hspace{-0.4cm}if } y_{ij} = 1, \\
        (1 - y_{ij})^\beta (p_{ij})^\alpha \log(1 - p_{ij}), & \text{\hspace{-0.3cm}otherwise}
        \end{cases}
\end{align}
\endgroup
with \( p_{ij} \) and \( y_{ij} \) being the pixel values at \((i, j)\) of the predicted \( (H_{\text{pred}}) \) and ground truth \( (H_{\text{gt}}) \) heatmaps, respectively; \( \{\alpha,\beta\} = \{2,4\} \) balance the loss terms; \( \mathcal{S} \) is number of pixels with \( y_{ij} = 1 \) (\( \mathcal{S} = 1 \) if none exist).
\( H_{\text{gt}} \) is created using unnormalized Gaussian probability density functions centered at each keypoint with a 2-pixel standard deviation.

The overall training loss is a weighted combination of the two losses: \(\mathcal{L} = \mathcal{L}_{joint} + \alpha(t) \mathcal{L}_{kp}.\) where \( \alpha(t) \), dependent on epoch \( t \), balances their relative importance. Since the joint angles are predicted in radians, \( \mathcal{L}_{joint} \) tends to be much smaller than \( \mathcal{L}_{kp} \), especially in early training. To address this, \( \alpha(t) \) is initialized at 0.0001, increased to 0.01 after 5 epochs, 0.1 after 10 epochs, and finally to 1 after 40 epochs, ensuring a balanced curriculum for training.

\subsection{Robot Pose Estimation}
\label{sec:pose_estimation}

\begin{figure}
    \centering
    \includegraphics[width=0.9\linewidth]{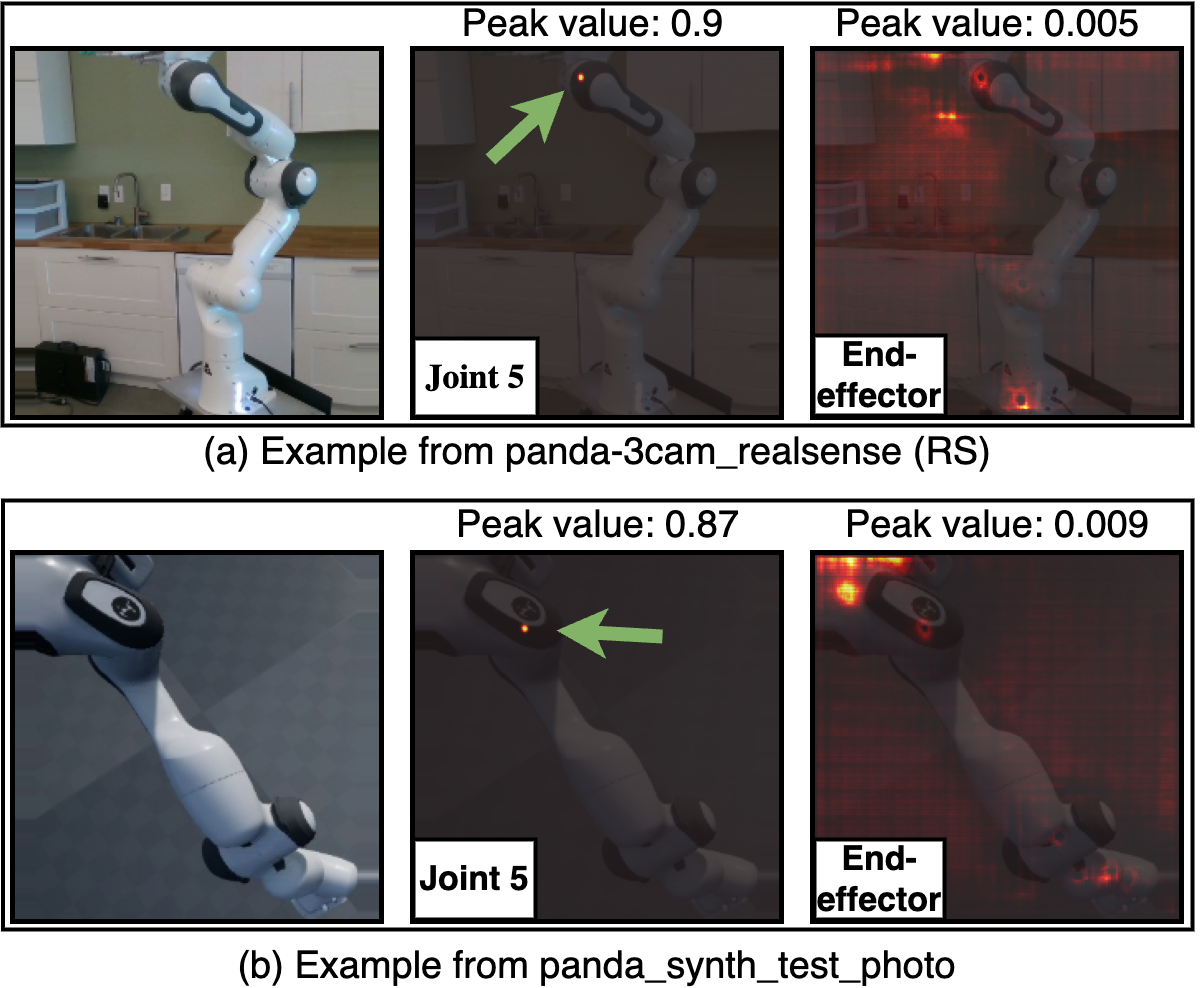}
    \caption{The examples show predicted heatmaps for Joint 5 and the End-Effector overlaid on the original image. The End-Effector, being positioned outside the field of view, produces noisy heatmaps with lower confidence (measured by peak values). Heatmap pixel values are normalized for better visualization. The green arrows highlight the peak values for Joint 5 for visual clarity.}
    \label{fig:kp_filtering_fig}
    \vspace*{-0.5cm}
\end{figure}

\noindent \textbf{Keypoint Filtering}:
The final layer of the Keypoint Net contains a sigmoid nonlinearity, that produces heatmaps with pixel values between 0 and 1, representing keypoint confidence at each pixel. The pixel with the highest confidence indicates the keypoint location.
However, when only a portion of the robot is visible, some keypoints may lie outside the image, leading to low confidence scores across the heatmap for these keypoints (\cref{fig:kp_filtering_fig}).
Selecting the pixel with the highest confidence in such cases can be misleading, as no pixel accurately represent the true keypoint. 
To address this, during evaluation, we apply a threshold \( \epsilon \), only considering keypoints with confidence above it.
For use with a PnP algorithm~\cite{epnp} for pose estimation, we require a minimum of four 2D-3D correspondences. If fewer than four keypoints remain after filtering, we iteratively reduce \( \epsilon \) by 0.025 until at least four keypoints are retained.

\noindent \textbf{Pose Estimation}:
The robot's pose is estimated using the EPnP algorithm~\cite{epnp} with the filtered 2D-3D correspondences and known camera intrinsics. 
As keypoints are defined on joints, we obtain the 3D points corresponding to the 2D keypoints using the robot's forward kinematics and predicted joint angles.

\noindent \textbf{Sim-to-Real Self-Supervised Training}:
In addition to supervised training for pose estimation, our method supports self-supervised fine-tuning of the trained models on real-world data to bridge the sim-to-real gap. 
Specifically, we use a differentiable PnP algorithm~\cite{ctrnet, differentiable_pnp} and estimate the robot's pose and transform 3D joint locations from the robot to the camera frame. These transformed points are projected onto the image plane, yielding the projected keypoints, \( P_{proj} \in \mathbb{R}^{k \times 2}\). We then minimize the mean squared error between \( P_{proj} \) and the predicted keypoints \( P_{pred} \)
\vspace*{-0.3cm}
\begin{align}
    \mathcal{L}_{\text{ssl}} = \frac{1}{k} \sum_{i=1}^k \| P_{pred}^i - P_{proj}^i \|_2^2
\end{align}
where $P_{pred}^i$ and $P_{proj}^i$ are the $i^{th}$ keypoint in $P_{pred}$ and $P_{proj}$, respectively.
During this sim-to-real finetuning, we use a lower learning rate than in the prior supervised training.
To prevent model collapse, the Keypoint Net's learning rate is set close to zero. Further details on learning rates are provided in the supplementary material.

\begin{table*}[h!]
\centering
\footnotesize
\setlength{\arrayrulewidth}{0.3mm} 
\begin{tabular}{l c c c c c c c c c c c}
\hline
\multirow{4}{*}{Method} & \multirow{4}{*}{\shortstack{Known \\ Joint\\Angles}} & \multirow{4}{*}{\shortstack{Known \\Bounding \\ Box}} & \multicolumn{6}{c}{Panda} & \multicolumn{2}{c}{Kuka} & Baxter \\

\cmidrule(lr){4-9} \cmidrule(lr){10-11} \cmidrule(lr){12-12}
 
 &  &  & \multicolumn{2}{c}{Synthetic} & \multicolumn{4}{c}{Real} & \multicolumn{2}{c}{Synthetic} & Synthetic \\
 \cmidrule(lr){4-5} \cmidrule(lr){6-9} \cmidrule(lr){10-11} \cmidrule(lr){12-12}
 
 &  &  & Photo & DR & AK & XK & RS & ORB & Photo & DR & DR \\

\hline

DREAM-F & \cellcolor{red!20}Yes & \cellcolor{green!20}No & 79.5 & 81.3 & 68.9 & 24.4 & 76.1 & 61.9 & - & - & - \\

DREAM-Q & \cellcolor{red!20}Yes & \cellcolor{green!20}No & 74.3 & 77.8 & 52.4 & 37.5 & 78.0& 57.1 & - & - & 75.5 \\

DREAM-H & \cellcolor{red!20}Yes & \cellcolor{green!20}No & 81.1 & 82.9 & 60.5 & 64.0& 78.8 & 69.1 & 72.1 & 73.3 & - \\

\hline

HPE & \cellcolor{green!20}No & \cellcolor{red!20}Yes & 82.0& 82.7 & 82.2 & 76.0& 75.2 & 75.2 & 73.9 & 75.1 & 58.8 \\

\hline

RoboPose & \cellcolor{green!20}No & \cellcolor{green!20}No & 79.7 & 82.9 & 70.4 & 77.6 & 74.3 & 70.4 & 73.2 & \cellcolor{gray!20}\textbf{80.2}& 32.7 \\

HPE$^*$ & \cellcolor{green!20}No & \cellcolor{green!20}No & 40.7 & 41.4 & 66.7 & - & 49.1 & 51.6 & 56.7 & 56.2 &  9.8\\

\textbf{RoboPEPP (Ours)} & \cellcolor{green!20}No & \cellcolor{green!20}No & \cellcolor{gray!20}\textbf{84.1}& \cellcolor{gray!20}\textbf{83.0}& \cellcolor{gray!20}\textbf{75.3}& \cellcolor{gray!20}\textbf{78.5}& \cellcolor{gray!20}\textbf{80.5}& \cellcolor{gray!20}\textbf{77.5}& \cellcolor{gray!20}\textbf{76.1}& 76.2& \cellcolor{gray!20}\textbf{34.4}\\

\hline

\end{tabular}
\caption{Comparison of robot pose estimation using AUC on the ADD metric. Best values among methods using unknown joint angles and bounding boxes during evaluation are bolded. HPE$^*$ denotes HPE~\cite{hpe} evaluated with the same off-the-shelf bounding box detector as RoboPEPP. HPE$^*$ was not evaluated on Panda XK since corresponding model weights were unavailable.}
\label{tab:auc_add}
\vspace*{-0.4cm}
\end{table*}

\noindent \textbf{Region of Interest (RoI) Detection}: 
During evaluation, we utilize the GroundingDINO~\cite{grounding_dino} object detection model to automatically locate the region of interest around the robot.  The detected region is cropped and resized to \( 224 \times 224 \) pixels. Further details are provided in the supplementary material.
During training, we use ground truth bounding box information. However, to ensure our model's robustness to region-of-interest detection, we employ a training curriculum: we use the ground truth bounding boxes and expand them by adding random offsets sampled from the uniform distribution \(\mathcal{U}(0, \lambda)\) to their edges, with \(\lambda\) progressing from 0 (first 30 epochs) to 30 pixels at epoch 30, 50 pixels at epoch 50, 80 pixels at epoch 70, 100 pixels at epoch 90, and 120 pixels at epoch 110. This approach encourages the model to generalize effectively, even with noisy region-of-interest detection during inference.

%% file: sec/4_experiments.tex
\section{Experiments}
\label{sec:experiments}
\subsection{Dataset and Implementation Details}
We evaluate our framework on the DREAM dataset~\cite{dream} that includes three robots (Franka Emika Panda, Kuka iiwa7, Rethink Baxter) and contains the following for each robot:  Panda  -- synthetic domain-randomized (DR) training, DR test (Panda DR), photo-realistic test (Panda Photo), four real-world test (Panda AK, XK, RS, ORB) sequences; Kuka -- synthetic DR training, DR test (Kuka DR), photo-realistic test (Kuka Photo) sequences; Baxter -- synthetic DR training and test (Baxter DR) sequences.

The encoder is pre-trained using our self-supervised embedding predictive strategy (\cref{sec:pre-training}) for 200 epochs on the DR training sequences of all robots, using AdamW~\cite{adamw} optimizer with an initial learning rate of $10^{-3}$. For end-to-end fine-tuning (\cref{sec:kp_joint_estimation}), models are trained separately for each robot for 200 epochs with AdamW optimizer (learning rate $10^{-4}$).
Sim-to-real fine-tuning is performed for 10 epochs. More details are in the supplementary material.

\subsection{Results}
\subsubsection{Robot Pose Prediction}
\label{sec:pose_prediction_results}
We evaluate RoboPEPP by computing the average distance~\cite{hpe,dream,robopose}, \(ADD = \frac{1}{n} \sum_0^n\|\hat{T}_R^C\hat{p}_i^{\{R\}} - p_i^{\{C\}} \|_2\) between predicted and ground truth joint positions in the camera frame for each image,
where $\hat{T}_R^C$ is the predicted robot pose in the camera frame, $\hat{p}_i^{\{R\}}$ is the estimated position of joint $i$ in the robot's frame (computed from predicted joints and forward kinematics), and $p_i^{\{C\}}$ is the ground truth joint position in the camera frame ($i=0$ signifies the robot base).

In Table \ref{tab:auc_add}, we report the area-under-the-curve (AUC) of the average distance (ADD) across various thresholds, where higher AUC values indicate greater accuracy. We compare RoboPEPP with Real-Time Holistic Robot Pose Estimation with Unknown States (referred to as HPE in this manuscript)~\cite{hpe}, RoboPose~\cite{robopose}, and three variants of DREAM~\cite{dream}, though the latter assume known joint angles. 
To the best of our knowledge, RoboPose, HPE, and RoboKeyGen~\cite{robokeygen} are the only approaches besides RoboPEPP that predict robot pose with unknown joint angles. However, RoboKeyGen evaluates on a different dataset, and its code is unavailable, preventing direct comparison. Nonetheless, its reported AUC on similar datasets is lower than ours.
Moreover, HPE~\cite{hpe} assumes known bounding boxes during evaluation, a condition often unrealistic in practice. Therefore, we also evaluate HPE with our bounding box detection strategy (denoted HPE$^*$) in Table \ref{tab:auc_add}. 

\begin{figure*}
    \centering
    \includegraphics[width=0.75\linewidth]{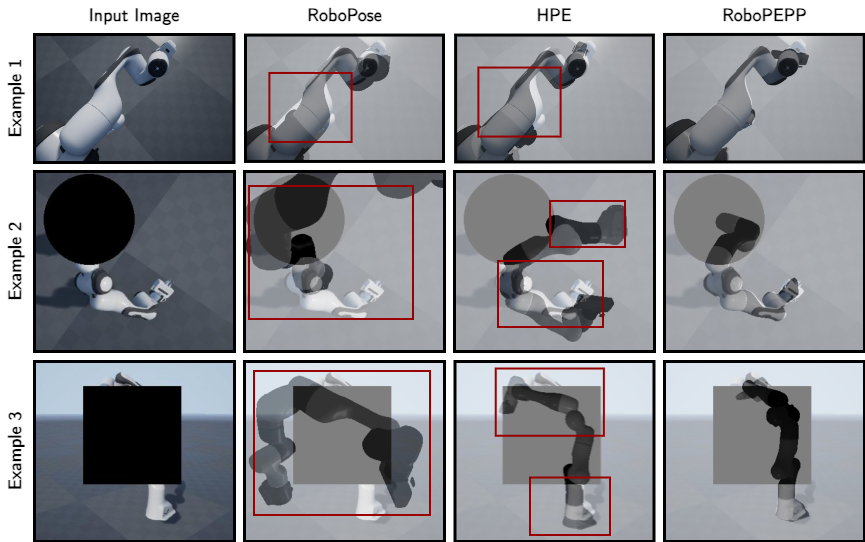}
    \caption{\textbf{Qualitative Comparison on Panda Photo (Example 1) and Occlusion (Example 2 and 3) datasets}: Predicted poses and joint angles are used to generate a mesh overlaid on the original image, where closer alignment indicates greater accuracy. Highlighted rectangles indicate regions where other methods' meshes misalign, while RoboPEPP achieves high precision.}
    \label{fig:comp_res}
    \vspace*{-0.5cm}
\end{figure*}

RoboPEPP yields the highest scores across all sequences (except for Kuka DR where it remains competitive) among methods with unknown joint angles and bounding boxes. HPE~\cite{hpe}, on the other hand, shows sensitivity to bounding box selection, with its performance dropping when ground truth bounding boxes are unavailable. 
Our analysis has shown that using bounding boxes just 5 pixels wider than ground truth reduces HPE's accuracy by up to 25\% on the Panda Photo test set and by around 50\% with 10-pixel wider boxes.
While both RoboPEPP and HPE are trained using ground truth bounding boxes, RoboPEPP's training strategy (\cref{sec:pose_estimation}) reduces dependency on them. Notably, RoboPEPP outperforms HPE on most sequences even when HPE uses known bounding boxes during evaluation.

A qualitative comparison of RoboPEPP with RoboPose~\cite{robopose} and HPE~\cite{hpe} on Panda Photo test dataset (example 1) and the occlusion dataset of \cref{sec:occlusion} (examples 2 and 3) is presented in \cref{fig:comp_res}. Each method uses the input image to predict pose and joint angles, rendering a robot mesh that is projected and overlaid onto the original image, where closer alignment indicates higher prediction accuracy. Example 1 depicts a case where only part of the robot is visible and examples 2 and 3 show cases of occlusions (detailed in \cref{sec:occlusion}). In these challenging scenarios, RoboPEPP achieves highly accurate overlays while other methods are less precise, as highlighted by the red rectangles.
In \cref{fig:comp_res}, HPE is used with ground truth bounding boxes.


\subsubsection{Joint Prediction}
In Table \ref{tab:joints}, we report the mean absolute error (in degrees) for joint angle prediction on the Panda Photo, Panda DR, Kuka Photo, and Kuka DR test datasets. The keypoints corresponding to the end-effector and the final joint (i.e., the joint nearest to the end-effector) lie along the axis of rotation of this joint, making their locations independent of this joint's angle. Consequently, we predict the angles of all joints except the last one, assigning a random angle to it during evaluation. Thus, Table \ref{tab:joints} presents the mean absolute error for the first six joint angles. 
We compare RoboPEPP with HPE~\cite{hpe} and RoboPose~\cite{robopose}. RoboPEPP demonstrates the lowest average joint prediction error on all datasets. Although HPE utilizes known bounding boxes for region-of-interest detection, RoboPEPP still outperforms HPE by over 15\% on average across all datasets in Table \ref{tab:joints}. When HPE is tested without ground truth bounding boxes, its performance drops to an average error of 7.2 degrees. 

\begin{figure}
    \centering
    \includegraphics[width=0.93\linewidth]{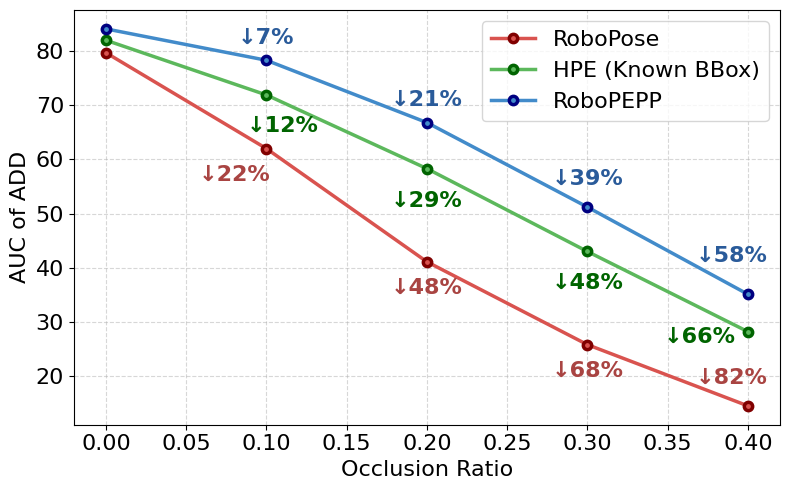}
    \caption{AUC comparison of the distance metric under varying occlusion levels, evaluated on the dataset in \cref{sec:occlusion}. Percentages next to the lines indicate the relative drop in each method's performance compared to their performance without occlusions.}
    \label{fig:occlusion}
    \vspace*{-0.4cm}
\end{figure}

\begin{table}
\centering
\footnotesize
\setlength{\tabcolsep}{2.5pt}
\begin{tabular}{l l |l |l l l l l l | l}
  \hline
  & & Method & J1 & J2 & J3 & J4 & J5 & J6 & Avg. \\
  \hline
  \multirow{6}{*}{\rotatebox{90}{Panda}} & \multirow{3}{*}{\rotatebox{90}{Photo}} & RoboPose & 7.7& 3.5& 4.3& 3.4& 7.3& 8.1& 5.7\\
  & & HPE (Known BBox)& 6.1& 2.2& 3.6& 2.0& 6.2& 6.6&   4.5\\
  & & \textbf{RoboPEPP} & \cellcolor{gray!0}\textbf{4.4}& \cellcolor{gray!0}\textbf{1.8}& \cellcolor{gray!0}\textbf{2.2}& \cellcolor{gray!0}\textbf{1.8}& \cellcolor{gray!0}\textbf{4.4}& \cellcolor{gray!0}\textbf{4.8}& \cellcolor{gray!0}\textbf{3.2}\\
    \cline{2-10}
  & \multirow{3}{*}{\rotatebox{90}{DR}} & RoboPose & 6.1& 2.7& 3.6& 2.5& 6.3& 8.1& 4.9\\
  & & HPE (Known BBox)& 6.2& \cellcolor{gray!0}\textbf{2.2}& 3.9& \cellcolor{gray!0}\textbf{1.9}& 5.9& 6.6&  4.4\\
  & & \textbf{RoboPEPP} & \cellcolor{gray!0}\textbf{4.9}& 2.3& \cellcolor{gray!0}\textbf{2.7}& 2.2& \cellcolor{gray!0}\textbf{4.9}& \cellcolor{gray!0}\textbf{5.4}& \cellcolor{gray!0}\textbf{3.8}\\
  \hline

  \multirow{6}{*}{\rotatebox{90}{Kuka}} & \multirow{3}{*}{\rotatebox{90}{Photo}} & RoboPose & 4.9& 5.1& 6.7& 6.0& 10.8& 9.6&   7.2\\
  & & HPE (Known BBox)& 4.8& 3.8& 5.0& \cellcolor{gray!0}\textbf{2.8}& 4.9& 5.9&   4.5\\
  & & \textbf{RoboPEPP} & \cellcolor{gray!0}\textbf{3.8}& \cellcolor{gray!0}\textbf{2.8}& \cellcolor{gray!0}\textbf{4.6}& 3.1& \cellcolor{gray!0}\textbf{3.8}& \cellcolor{gray!0}\textbf{5.4}&  \cellcolor{gray!0}\textbf{3.9}\\
  \cline{2-10}
   & \multirow{3}{*}{\rotatebox{90}{DR}} & RoboPose & 4.4& \textbf{2.8}& 5.4& 3.4& 12.5& 8.5&  6.2\\
  & & HPE (Known BBox)& 4.6& 3.6& \cellcolor{gray!0}\textbf{4.9}& \cellcolor{gray!0}\textbf{2.8}& 5.2& \cellcolor{gray!0}\textbf{6.1}&  4.5\\
  & & \textbf{RoboPEPP} & \cellcolor{gray!0}\textbf{3.7}& 3.5& 5.1& 3.5& \cellcolor{gray!0}\textbf{4.1}& 6.2&  \cellcolor{gray!0}\textbf{4.3}\\
  \hline
   & \multirow{3}{*}{\rotatebox{90}{Avg.}} & RoboPose & 5.8& 3.5& 5.0& 3.8& 9.2& 8.6&  6.0\\
  & & HPE (Known BBox)& 5.4& 3.0& 4.4& \cellcolor{gray!20}\textbf{2.4}& 5.6& 6.3&  4.5\\
  & & \textbf{RoboPEPP} & \cellcolor{gray!20}\textbf{4.2} & \cellcolor{gray!20}\textbf{2.6} & \cellcolor{gray!20}\textbf{3.7} & 2.7 & \cellcolor{gray!20}\textbf{4.3} & \cellcolor{gray!20}\textbf{5.5}&  \cellcolor{gray!20}\textbf{3.8}\\
 
  \hline
\end{tabular}
\caption{Mean absolute error between the predicted and actual joint angles (in degrees) for the Panda and Kuka synthetic test sets. }
\label{tab:joints}
\vspace*{-0.5cm}
\end{table}

\begin{figure}
    \centering
    \includegraphics[width=0.8\linewidth]{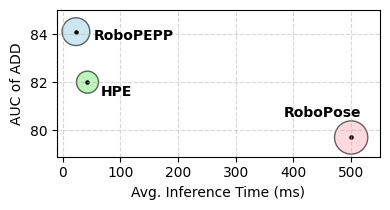}
    \caption{Execution time and computation analysis on the Panda Photo test dataset with RoboPEPP showing best performance and accuracy. The circle sizes in the plot represent model FLOPs.}
    \label{fig:exec_time}
    \vspace*{-0.6cm}
\end{figure}

\begin{figure*}[ht]
    \centering
    \begin{subfigure}[b]{0.4\linewidth}
        \centering
        \includegraphics[width=\linewidth]{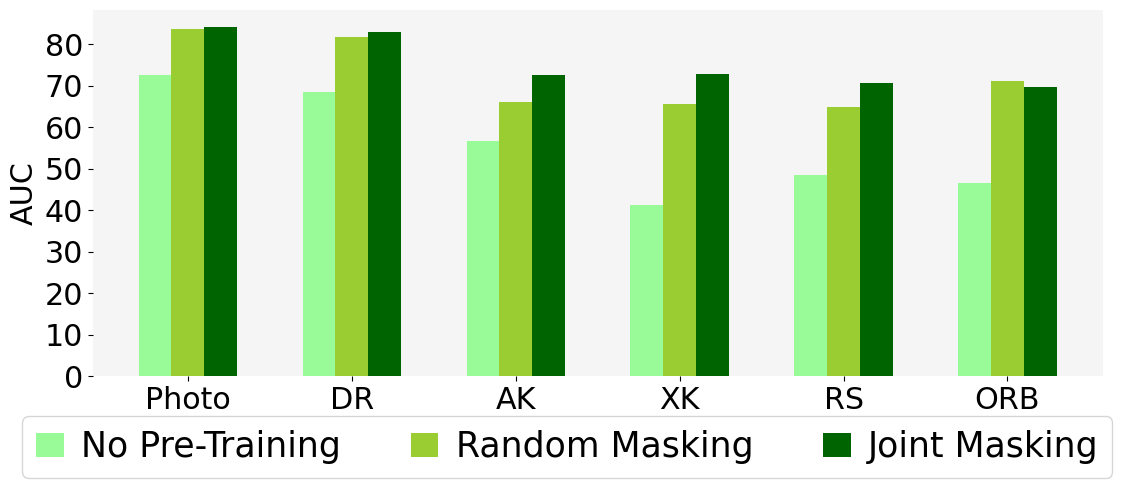}
        \caption{RoboPEPP's joint-masking-based pre-training achieves the best overall performance in robot pose estimation.}
        \label{fig:pre_training_results}
        \vspace*{-0.3cm}
    \end{subfigure}
    \hfill
    \begin{subfigure}[b]{0.28\linewidth}
        \centering
        \includegraphics[width=\linewidth]{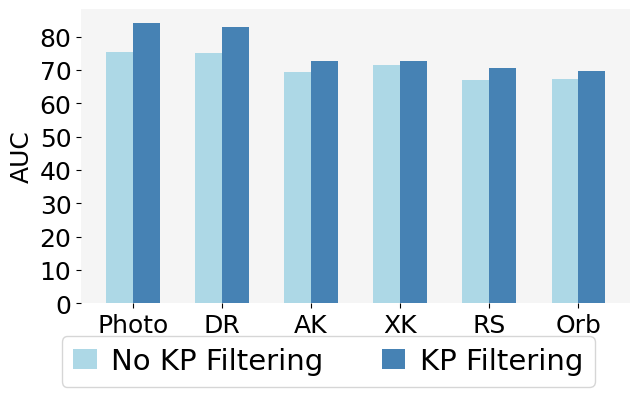}
        \caption{Keypoint Filtering leads to improved performance on all datasets.}
        \label{fig:kp_filtering_results}
        \vspace*{-0.3cm}
    \end{subfigure}
    \hfill
    \begin{subfigure}[b]{0.29\linewidth}
        \centering
        \includegraphics[width=\linewidth]{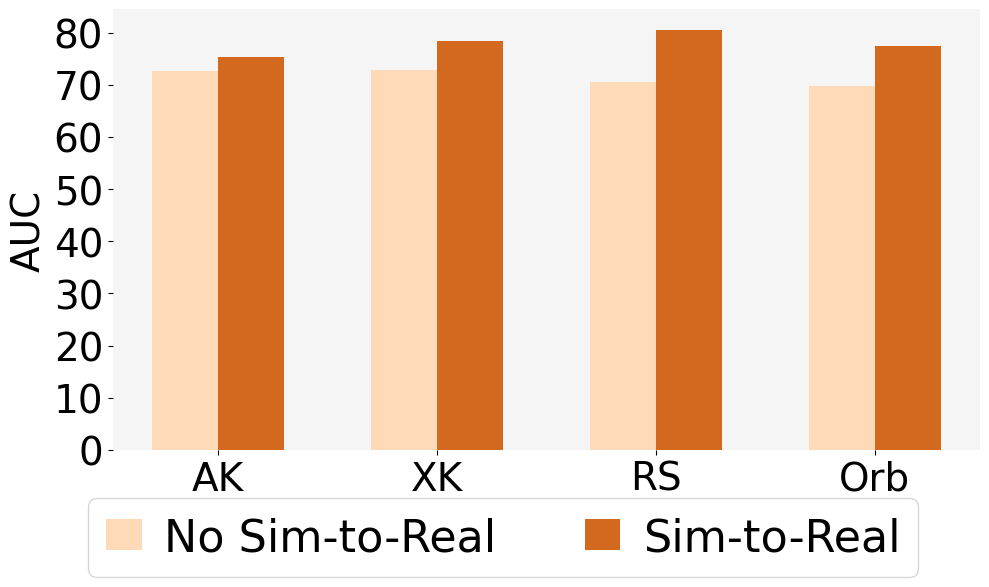}
        \caption{RoboPEPP's sim-to-real fine-tuning improves performance on real datasets.}
        \label{fig:sim2real_results}
        \vspace*{-0.3cm}
    \end{subfigure}
    \caption{Ablation studies on (a) Pre-Training, (b) Keypoint Filtering, and (c) Sim-to-Real fine-tuning on the Panda test datasets.}
    \label{fig:combined_results}
    \vspace*{-0.5cm}
\end{figure*}

\subsubsection{Robustness to Occlusions}  
\label{sec:occlusion}
In addition to achieving high performance across various metrics and datasets, RoboPEPP demonstrates robustness to occlusions. To evaluate this, we compared the performance of RoboPEPP with other methods~\cite{robopose, hpe} on a custom dataset, created by adding synthetic occlusions to Panda Photo. Specifically, we overlaid black rectangular or circular masks at random positions on the robot, ensuring that the masks covered at least some part of the robot (and not just the background).
We generated four test sequences with occlusion ratios of 0.1, 0.2, 0.3, and 0.4 on the RoI area in each image, respectively. This approach differs from the masking used during our training, where the model is informed about the number of masked patches. Here, the model processes occluded images as it would any other input image, without any knowledge of the occlusion.

In \cref{fig:occlusion}, we plot the AUC of the ADD against the occlusion ratio. The plot also includes the percentage decrease in AUC relative to the respective model's performance without occlusion. RoboPEPP demonstrates superior robustness to occlusion, achieving an AUC score of 35.1 even when 40\% of the RoI is occluded, compared to 28.2 for HPE and 14.5 for RoboPose.
Examples 2 and 3 in \cref{fig:comp_res} provide qualitative comparisons of RoboPEPP, HPE~\cite{hpe}, and RoboPose~\cite{robopose} on the occlusion dataset. 
RoboPEPP demonstrates superior performance, even in challenging cases like example 3, where most of the robot is occluded. In both examples 2 and 3, RoboPose produces inaccurate results, while HPE shows partial overlap but still exhibits notable inaccuracies, highlighted by the red rectangles.

\subsubsection{Percentage of Correct Keypoints}
The accuracy of 2D keypoint detection affects the overall performance of RoboPEPP. Therefore, in Table \ref{tab:pck}, we report the percentage of correct keypoints (PCK) within thresholds of 2.5, 5, and 10 pixels on the Panda Photo dataset. Since HPE~\cite{hpe} and RoboPose~\cite{robopose} do not rely on 2D keypoint detection for pose estimation, we include only DREAM~\cite{dream} as a baseline for comparison. 
RoboPEPP achieves high average PCK scores, with 0.43 @ 2.5 pixels, 0.73 @ 5 pixels, and 0.95 @ 10 pixels. 
While DREAM surpasses RoboPEPP in PCK@2.5 pixels on AK and XK, RoboPEPP achieves higher overall accuracy across most metrics, showcasing its robust keypoint detection.
We hypothesize that DREAM's known joint angles assumption allows it to focus entirely on keypoint detection enhancing PCK in some cases.

\begin{table}
\centering
\footnotesize
\setlength{\tabcolsep}{9.5pt}
\begin{tabular}{ l  l  c  c c }
\hline
\multirow{2}{*}{Dataset} & \multirow{2}{*}{Method} & \multicolumn{3}{c}{PCK @ (pixel)} \\
 &  & 2.5 & 5 & 10 \\
 \hline
\multirow{2}{*}{DR} & DREAM & 0.79 & 0.88 & 0.90 \\
 & \textbf{RoboPEPP} & \cellcolor{gray!0}\textbf{0.84} & \cellcolor{gray!0}\textbf{0.91} & \cellcolor{gray!0}\textbf{0.93} \\
  \hline
\multirow{2}{*}{Photo} & DREAM & 0.77 & 0.87 & 0.90 \\
 & \textbf{RoboPEPP} & \cellcolor{gray!0}\textbf{0.87} & \cellcolor{gray!0}\textbf{0.92} & \cellcolor{gray!0}\textbf{0.94} \\
  \hline
\multirow{2}{*}{AK} & DREAM & \textbf{0.36} & \textbf{0.65} & 0.90 \\
 & \textbf{RoboPEPP} & 0.16 & 0.62 & \textbf{0.93} \\
  \hline
\multirow{2}{*}{XK} & DREAM & \textbf{0.15} & \textbf{0.37} & 0.59 \\
 & \textbf{RoboPEPP} & 0.09 & \cellcolor{gray!0}\textbf{0.37} & \cellcolor{gray!0}\textbf{0.96} \\
  \hline
\multirow{2}{*}{RS} & DREAM & 0.24 & \textbf{0.83} & 0.96 \\
 & \textbf{RoboPEPP} & \cellcolor{gray!0}\textbf{0.31} & 0.82 & \cellcolor{gray!0}\textbf{0.97} \\
  \hline
\multirow{2}{*}{ORB} & DREAM & \textbf{0.28} & 0.67 & 0.83 \\
 & \textbf{RoboPEPP} & \cellcolor{gray!0}\textbf{0.28} & \cellcolor{gray!0}\textbf{0.73} & \cellcolor{gray!0}\textbf{0.96} \\
  \hline
\multirow{2}{*}{\textbf{Avg.}} & DREAM & \cellcolor{gray!20}\textbf{0.43} & 0.71 & 0.85 \\
 & \textbf{RoboPEPP} & \cellcolor{gray!20}\textbf{0.43} & \cellcolor{gray!20}\textbf{0.73} & \cellcolor{gray!20}\textbf{0.95} \\
  \hline

\end{tabular}
\caption{Comparison of Percentage of Correct Keypoints (PCK) at different pixel thresholds across the Panda test datasets.}
\label{tab:pck}
\vspace*{-0.4cm}
\end{table}

\subsubsection{Execution Time}
To demonstrate the practical effectiveness of the proposed RoboPEPP method, we compare execution times (in milliseconds) in \cref{fig:exec_time}. The circle sizes in the figure correspond to the relative number of floating-point operations (FLOPs) required by each model. All evaluations were conducted on a system equipped with an Nvidia RTX A4000 GPU, an Intel(R) i9 CPU, and 128 GB RAM, using the Panda Photo test dataset. Consistent with previous work~\cite{hpe}, we report only model execution time, excluding pre-processing steps such as data loading and RoI detection. Despite having a slightly higher FLOPs count than HPE~\cite{hpe}, RoboPEPP achieves the highest AUC ADD score and the fastest execution time, completing inference in just 23 milliseconds.

\subsection{Ablation Studies}
\noindent\textbf{Embedding Predictive Pre-Training:}
To assess the impact of embedding predictive pre-training in RoboPEPP, we conducted an ablation study comparing three versions of the model: a version of RoboPEPP without pre-training, and a version pre-trained with random masking instead of joint-specific masking, and standard RoboPEPP (i.e., pre-trained with joint masking). For all experiments, we utilized the same model architecture and training settings. The bar graphs in \cref{fig:pre_training_results} illustrate that pre-training significantly improves performance. While the model trained with the default masking strategy demonstrated competitive results on synthetic test datasets, the model trained with joint-specific masking achieved better performance on real-world datasets in general. 
Note that the real-world results shown here do not include the sim-to-real fine-tuning of \cref{sec:pose_estimation}.
Further, on the occlusion dataset (\cref{sec:occlusion}) with a 0.4 occlusion ratio,  the model without pre-training achieves AUC of 22.6, the one with random masking reaches 30, and RoboPEPP achieves 35.1, highlighting the latter's occlusion robustness.

\noindent \textbf{Keypoint Filtering}:
In \cref{fig:kp_filtering_results}, we demonstrate that the integration of keypoint filtering (KP filtering) enhances performance across all datasets by helping in filtering out keypoints that fall outside the camera's field of view. Similar to \cref{fig:pre_training_results}, the real-world results presented in \cref{fig:kp_filtering_results} do not include any sim-to-real fine-tuning.

\noindent \textbf{Sim-to-Real Fine-Tuning}:
In \cref{fig:sim2real_results}, we show performance gains from sim-to-real self-supervised training, with the model's accuracy improving by an average of $\sim$6 points after fine-tuning. Our sim-to-real training requires only 10 epochs, with each epoch lasting around 2 minutes.

%% file: sec/5_conclusion.tex
\section{Conclusion}
\label{sec:conclusion}
We introduced a novel framework RoboPEPP enhancing robot pose and joint angle estimation using an embedding predictive pre-training strategy. RoboPEPP uses a joint-masking-based method to pre-train an encoder-predictor pair to be integrated into downstream networks for joint and pose predictions. Experimental results show RoboPEPP's superior performance, particularly in handling occlusions due to the combined effects of pre-training and random masking during fine-tuning. RoboPEPP's training helps fuse knowledge of the robot's physical model within the encoder, making RoboPEPP effective for pose estimation and versatile for broader applications such as system dynamic prediction and 
imitation learning.